\documentclass[runningheads]{llncs}
\usepackage{cite} 
\usepackage{multirow} 
\usepackage{amsfonts}
\usepackage[T1]{fontenc}

\usepackage[acronym,toc,xindy]{glossaries} 
\newacronym{sam2}{SAM2}{Segment Anything 2}
\newacronym{sam}{SAM}{Segment Anything}
\newacronym{ilp}{ILP}{integer linear programming}
\newacronym{ctc}{CTC}{Cell Tracking Challenge}
\newacronym{aogm}{AOGM}{Acyclic Oriented Graph Matching}
\usepackage{graphicx,verbatim}
\usepackage{amsmath}
\usepackage{hyperref}
\usepackage{color}
\usepackage{pifont}

\begin{document}
\title{Segment Anything for Cell Tracking}
%
\author{Zhu Chen(\ding{41})\orcidID{0009-0009-9847-7686} \and Mert Edgü\orcidID{0009-0006-5691-8277}\and \\Er Jin \and
Johannes Stegmaier\orcidID{0000-0003-4072-3759}}
\authorrunning{Z. Chen et al.}
%
\institute{Institute of Imaging and Computer Vision, RWTH Aachen University, Germany\\
Email: \email{zhu.chen@lfb.rwth-aachen.de}}

\maketitle              

\begin{abstract}

Tracking cells and detecting mitotic events in time-lapse microscopy image sequences is a crucial task in biomedical research. However, it remains highly challenging due to dividing objects, low signal-to-noise ratios, indistinct boundaries, dense clusters, and the visually similar appearance of individual cells. Existing deep learning-based methods rely on manually labeled datasets for training, which is both costly and time-consuming. Moreover, their generalizability to unseen datasets remains limited due to the vast diversity of microscopy data. To overcome these limitations, we propose a \textbf{zero-shot} cell tracking framework by integrating \gls{sam2}, a large foundation model designed for general image and video segmentation, into the tracking pipeline. As a fully-unsupervised approach, our method does not depend on or inherit biases from any specific training dataset, allowing it to generalize across diverse microscopy datasets without fine-tuning. Our approach achieves competitive accuracy in both 2D and large-scale 3D time-lapse microscopy videos while eliminating the need for dataset-specific adaptation. The source code is publicly available at \href{https://github.com/zhuchen96/sam4celltracking}{https://github.com/zhuchen96/sam4celltracking}.


\end{abstract}
\section{Introduction}
Cell tracking plays a key role in understanding tissue formation, disease progression, embryonic development, and various biomedical applications~\cite{background3,background2}. The task includes tracking the movement and morphological changes of individual cells, identifying parent-daughter cell pairs during mitosis, detecting cell interactions, and reconstructing full lineage information~\cite{tracking1}.

Most current methods follow the tracking-by-detection paradigm, where objects are first detected in individual frames and then linked over time to form trajectories. The linking of segmented masks is typically performed as a post-processing step, which can be solved by simple approaches like nearest-neighbor matching. Alternatively, it can be formulated as a global optimization problem and solved by \gls{ilp}~\cite{ilp,ilp2} or the Viterbi algorithm\cite{viterbi}. In this case, the performance of tracking heavily depends on the quality of the segmentation results~\cite{ultrack}. Recently, deep learning-based methods have significantly improved both cell segmentation~\cite{cellpose,stardist,mediar} and tracking~\cite{gnn,trackastra,caliban,ce_dl,motion1,motion2}. In addition, by fine-tuning foundation models like \gls{sam}~\cite{sam} and \gls{sam2}~\cite{sam2} on diverse microscopy datasets, they demonstrate strong performance in 2D microscopy image segmentation~\cite{cellsam, sac}. MicroSAM~\cite{microsam} extends \gls{sam} for volumetric segmentation and tracking in microscopy data, but it relies on projecting 2D segmentation masks onto adjacent slices and time frames, which limits accuracy and structural analysis. Additionally, \cite{medsam2} and \cite{sam2-medsam} treat different slices of 3D image data as a video sequence, allowing \gls{sam2} to be applied for volumetric segmentation. Among these methods, training with datasets spanning multiple imaging modalities and cell types has been shown to enhance generalizability. However, microscopy image analysis faces a major challenge due to the vast diversity of cell types and imaging techniques. Methods such as phase-contrast, fluorescence, and bright-field microscopy each have unique characteristics~\cite{ctc2}. As a result, existing methods often fail to maintain accuracy when applied to a completely different type of data.

To address the challenge, we propose a novel \textbf{zero-shot} 2D and 3D cell tracking pipeline by integrating the pretrained \gls{sam2} model. As a fully-unsupervised approach, our method avoids dataset-specific biases and removes the requirement for manual segmentation and tracking annotations. 

For 2D and small 3D datasets, we employ a linking-only approach, where we construct a complete lineage graph from a pre-segmented set of masks and correct any failed detections. 

For large-scale 3D datasets, where detection and segmentation are significantly more challenging, we simultaneously address tracking and segmentation by incorporating SAM-Med3D~\cite{sammed3d}, a medical foundation model trained without microscopy images. By fine-tuning it on rough segmentation masks generated by simple methods such as watershed, we enable the model to produce accurate 3D segmentation results during the tracking process. Compared to global optimization-based approaches, our tracking pipeline allows for the targeted tracking of specific subsets of cells, reducing the influence of blurring and noisy areas in large-scale images. Our key contributions are as follows:

\begin{enumerate}
\item With appropriate prompt selection, the pre-trained \gls{sam2} model can be applied to segment and link cells based on previous mask information in microscopy image sequences without requiring annotated data.
\item By integrating SAM-Med3D, our pipeline achieves volumetric segmentation and tracking performance comparable to fully-supervised algorithms.
\end{enumerate}

\section{Method}

\subsection{Cell Linking in 2D and 3D Time-lapse Microscopy Images} \label{2dlinking}

\begin{figure}
\includegraphics[width=\textwidth]{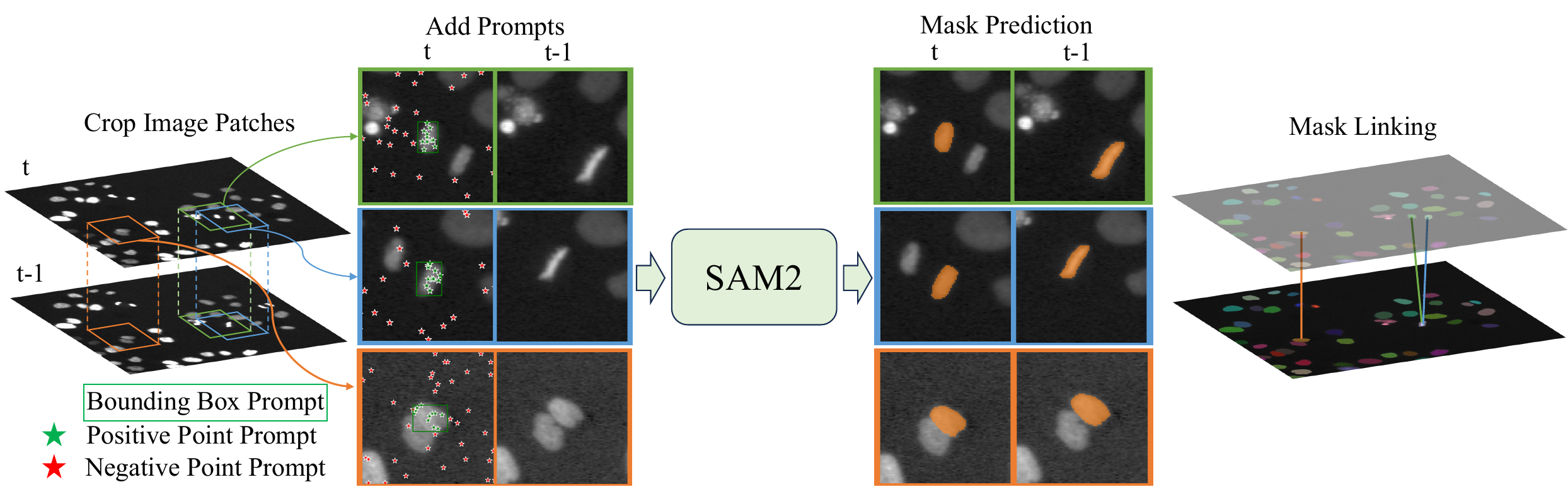}
\caption{Overview of the cell linking method. Patches are cropped from consecutive images based on the mask center at time point $t$. Bounding box and point prompts are then generated from the mask at time $t$. \gls{sam2} predicts the segmentation masks of the same cell in both time frames, and the masks are linked to form tracklets. The depicted example is the Fluo-N2DL-HeLa dataset from Cell Tracking Challenge (\gls{ctc})~\cite{ctc}.
} \label{fig:linking}
\end{figure}

Cell segmentation masks are typically generated independently for each time frame and must be linked to reconstruct full lineage trees, incorporating cell migration and mitotic information. For a sequence of $T$ raw images $I_1, I_2, \dots, I_T \in \mathbb{R}^{w \times h}$, each image $I_t$ has an associated segmentation mask set $M_t$, which can be generated using any segmentation approach. A cell lineage $L_k$ is defined as a temporally continuous sequence of segments: $L_k = \{m_t^{k,s} \mid t_1 \leq t \leq t_2, s \in M_t\}$, where each segmentation mask $m_t^{k,s}$ of tracklet $k$ corresponds to a specific segment $s$ at time $t$. Temporal consistency ensures that each time point within $[t_1, t_2]$ contains exactly one mask per lineage $L_k$.

We initialize tracking at the final time frame $T$, where a user interface allows refinement of detections (Suppl. Video Part 1), yielding an initial mask set $\tilde{M}_T$. The algorithm then propagates masks \textbf{backward} in time, linking across frames. For each mask $m_t^{k,s}$, the center $c_t^{k,s} = (x_t^{k,s}, y_t^{k,s})$ is computed, and a square patch $p_t^{k,s}$ of side length $d$ is extracted:
\begin{equation}
p_t^{k,s} = I_t \left[ x_t^{k,s} - \frac{d}{2} : x_t^{k,s} + \frac{d}{2}, \ y_t^{k,s} - \frac{d}{2} : y_t^{k,s} + \frac{d}{2} \right].
\end{equation}
The corresponding patch $p_{t-1}^{k,s}$ is extracted from $I_{t-1}$ at the same coordinates, as depicted in Fig.~\ref{fig:linking}. The pair $\{p_{t}^{k,s}, p_{t-1}^{k,s}\}$ is provided to \gls{sam2} as a short video sequence. Using the known mask \( m_t^{k,s} \), we generate the following prompts: 1. Bounding box $bb_t^{k,s}$ of $m_t^{k,s}$. 2. Positive points from $m_t^{k,s}$ and negative points from the background. By taking the input image patch sequence $\{p_{t}^{k,s}, p_{t-1}^{k,s}\}$ and applying the prompts on $p_{t}^{k,s}$, we apply the mask decoder to predict the potential mask for frame $t-1$. The potential mask is then compared to the pre-segmented masks in $M_{t-1}$. If a mask sufficiently overlaps, it is labeled as $m_{t-1}^{k,s}$ and linked to $L_k$. If the potential mask is too small or touches the image boundary, tracking is terminated; otherwise, if the predicted mask is valid but missing in \( M_{t-1} \), it is added to \( L_k \) and \( M_{t-1} \) as a recovered detection. Mitosis is detected when two masks in frame $t$ link to the same mask in $t-1$. If multiple masks link to one, only the two closest centers are retained. New tracklets are also initialized in intermediate time frames when unlinked masks remain in $M_{t-1}$. For 3D datasets, since the masks are volumetric, the image patches $\{p_{t}^{k,s}, p_{t-1}^{k,s}\}$ are extracted from the depth slice with largest mask area for each mask. Due to anisotropic voxel resolution, the x-y plane offers more reliable linking information.

\begin{figure}
\includegraphics[width=\textwidth]{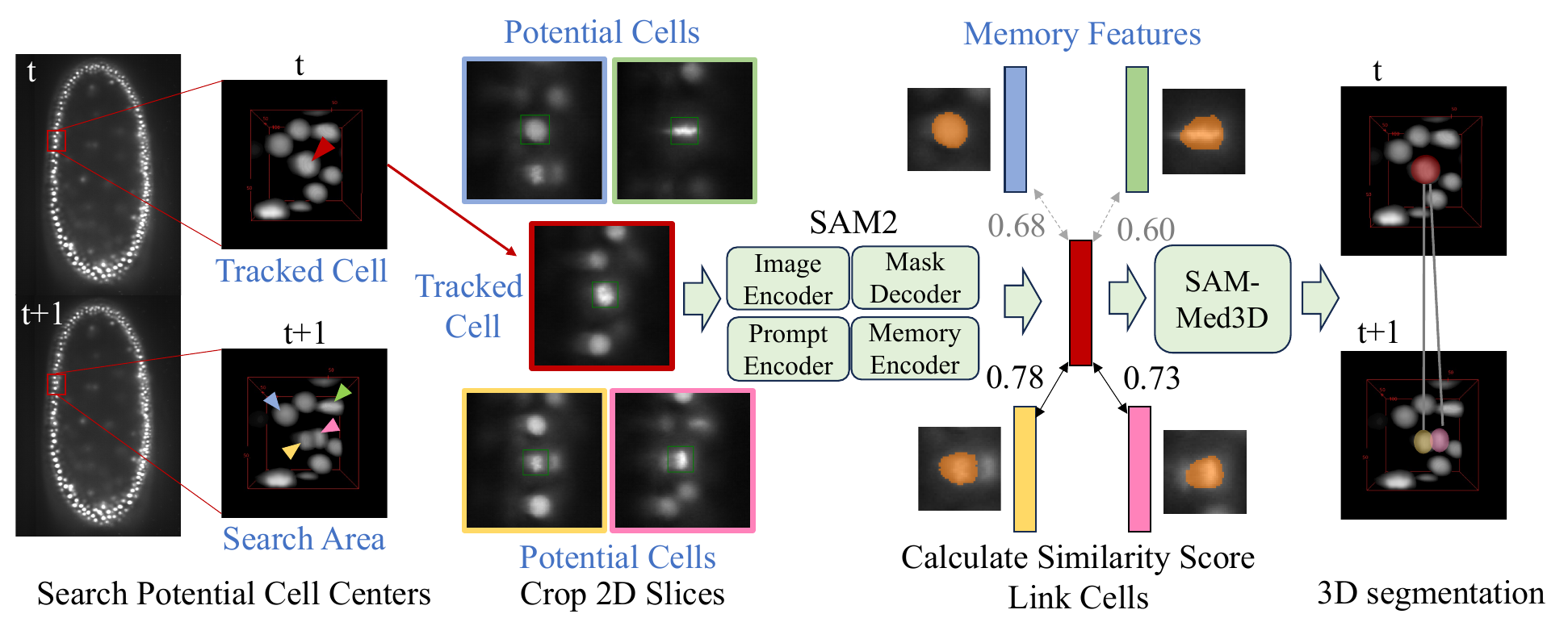}
\caption{For a given pair of consecutive time frames, potential cells in time point $t+1$ are searched within the neighborhood of the tracked cell in time point $t$. The bounding box of the tracked cell is used as a prompt for all local patches of potential cells. We use \gls{sam2} to extract memory feature vectors for all patches, and the cell with the highest similarity score is linked. A mitotic event is detected when two potential cells have close similarity scores, then the two daughter cells are linked to the same parent cell. In the last step, 3D masks are generated by SAM-Med3D. The depicted example is the Fluo-N3DL-TRIF dataset from CTC~\cite{ctc}.} \label{fig:tracking}
\end{figure}

\subsection{Cell Tracking in Large-scale 3D+t Microscopy Images}

Large-scale 3D time-lapse microscopy images can reach several gigabytes per time frame and contain up to thousands of cells. Pre-segmented masks often contain
missing or redundant detections, making global optimization-based linking inaccurate and inefficient for such large data. To address this, we develop a complete pipeline for simultaneous cell segmentation and tracking. 

First, we apply the method introduced in \cite{twang} to detect potential cell centers for each time frame $t$, and generate rough segmentation masks. We then fine-tune SAM-Med3D using these rough masks, enabling it to generate 3D segmentation masks with a single click on the cell area. To facilitate refinement and selection, we provide an interactive interface (see Suppl. Video Part 1), allowing users to adjust detections or choose a specific group of cells for tracking in the initial frame. 

Tracking is performed \textbf{forward} in time through each pair of consecutive time frames $t$ and $t+1$, as illustrated in Fig.~\ref{fig:tracking}. For each tracked cell at time $t$, we search for neighboring cell centers in time frame $t+1$ within a predefined radius $\tau$, designating them as potential candidates for linking. We then crop image patches centered on each candidate cell center and process these patches, along with the tracked cell's patch, through \gls{sam2}’s image encoder to extract local visual features. Using the tracked cell's mask, we compute its bounding box, which serves as a prompt for \gls{sam2} to generate a segmentation mask at the predicted location. The same bounding box prompt is also applied to all candidate patches to predict a potential mask of the same size as the tracked mask in the center of each patch. Then we take the high-level spatial features (the top-level feature map) from the image patches and the predicted segmentation mask and feed them to the memory encoder of \gls{sam2}. It encodes the memory representation that contains both the visual content of the local patch and the semantic information of the mask. To determine the correct match, we compute the cosine similarity between the memory-encoded features of the tracked cell at time $t$ and all candidate cells at time $t+1$. If a candidate cell surpasses a similarity threshold, it is linked to the lineage. In the presented experiments, we fixed this threshold to 0.8, but it might need adjustments in other circumstances. If no highly similar candidate is found, the two candidates with the highest similarity are compared. If the difference in their similarity scores is below 0.1, indicating they share local visual features but have different appearances, they are classified as daughter cells. In this case, a mitotic event is detected and two new tracklets are then initialized. If no suitable candidate is identified, the missing cell’s position is predicted using the method described in Section \ref{2dlinking}. Once the matching cell center is found in time point $t+1$, the fine-tuned SAM-Med3D is applied to generate the corresponding 3D segmentation mask.

This method not only introduces new central points for missed detections but also discards redundant ones with low similarity scores. Additionally, all images are stored in Zarr format \cite{zarr}, allowing the whole tracking pipeline—including the fine-tuning of SAM-Med3D—to work only with small local patches rather than entire large images, significantly reducing computational costs compared to global optimization-based approaches. Furthermore, since each lineage is processed independently, the method can be parallelized, further enhancing efficiency.
\section{Experiments}
\subsection{Metrics} 

As the evaluation metric for cell tracking accuracy, \gls{aogm}~\cite{aogm} is used to compare predicted and ground-truth cell lineage graphs. Each tracking result is represented as a graph, where nodes represent detected cells at specific time points, and directed edges denote migration or mitosis events. \gls{aogm} quantifies tracking errors by computing the graph edit distance with weighted penalties for false positives, false negatives, missing or incorrect links, and mitosis errors. The final AOGM score is the sum of these weighted costs, where a lower score indicates higher tracking accuracy. For normalized evaluation, the tracking accuracy measure (TRA) is defined as:
\begin{equation}
    \textrm{TRA} = 1 - \frac{\min(\textrm{AOGM},\textrm{AOGM}_0)}{\textrm{AOGM}_0},
\end{equation}
where \( \textrm{AOGM}_0 \) is the cost of constructing the reference graph from scratch. A TRA value close to 1 indicates better tracking accuracy. For large-scale 3D datasets, where segmentation masks are generated during tracking, we use the Jaccard-similarity-index-based segmentation accuracy measure (SEG) score to evaluate segmentation quality. It measures the Jaccard similarity index \( J \) between reference mask \( R \) and predicted mask \( S \):
\begin{equation}
    J(S, R) = \frac{|R \cap S|}{|R \cup S|}.
\end{equation}
A match is considered valid if the overlap exceeds 50\% of \( R \). The value range of SEG is between $[0, 1]$ and a higher value indicates better segmentation accuracy.

\begin{figure}
\includegraphics[width=\textwidth]{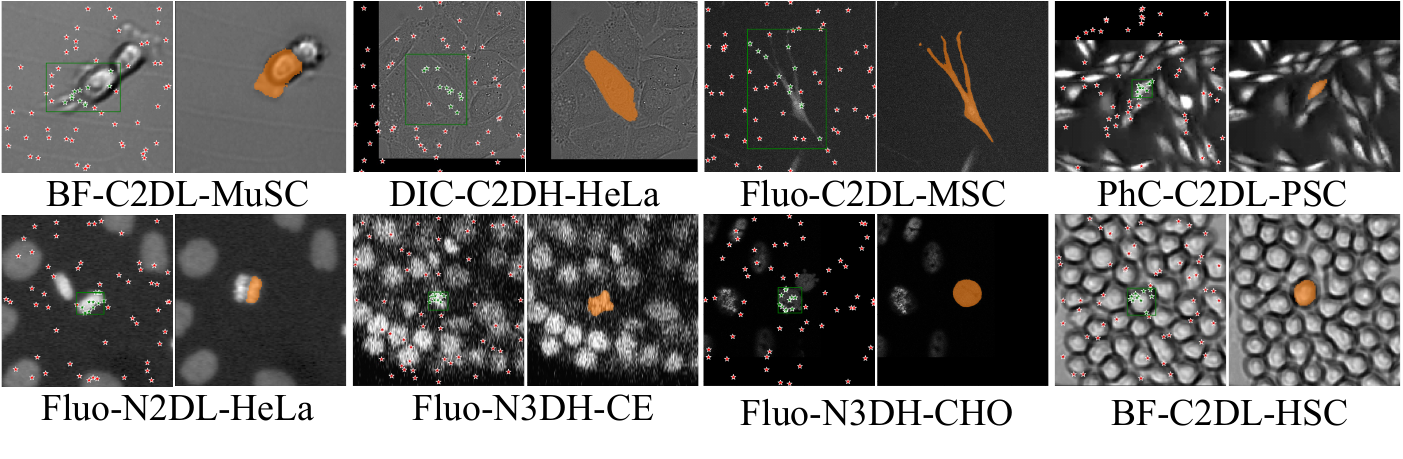}
\caption{Visualization of several examples of \gls{sam2} prediction results. Left: input image patch at time \( t \) with prompts. Right: output image patch at time \( t-1 \) with the predicted mask. The depicted examples are from CTC~\cite{ctc}.} \label{fig:visualize}
\end{figure}

\glsreset{ctc}
\subsection{Cell Linking Results}

We evaluate cell linking performance using 2D and 3D datasets from the \gls{ctc}~\cite{ctc,ctc2}. Each dataset provides an incomplete or complete set of segmentation masks, and our goal is to reconstruct the full lineage graph. The AOGM-based linking accuracy measure (LNK) compares the predicted lineage graph with manually labeled ground truth, while the biological accuracy measure (BIO) assesses tracking performance using biologically relevant metrics such as full lineage reconstruction, longest correctly tracked segments, mitosis detection, and cell cycle duration. Since our method is based on the pretrained \gls{sam2} model, no labeled training data for tracking is required. All parameters are selected based on the segmented cell size, without refinement with the ground truth data. Our approach is tested on 13 blind test datasets spanning different cell types and imaging techniques; the scores for selected datasets are provided in Tab.~\ref{tab1} and several examples of mask prediction are shown in Fig.~\ref{fig:visualize}. The hardware used in this work is an NVIDIA RTX 4090 (24 GB). We employed sam2.1\_hiera\_l as the pretrained SAM2 model. Tracking a sequence with 200 frames and about 100 cells per frame takes roughly one hour.

\begin{table}[ht]
\caption{Linking results on the selected datasets from \gls{ctc}~\cite{ctc}; the number in the superscript represents the corresponding method; our submission is in bold.}\label{tab1}
\centering
\begin{tabular}{cc cc cc cc}
\hline
\multicolumn{2}{c}{Dataset} & \multicolumn{2}{c}{All 13 Datasets} & \multicolumn{2}{c}{Fluo-N3DH-CE} & \multicolumn{2}{c}{Fluo-N3DH-CHO}  \\
\hline
\multicolumn{2}{c}{\multirow{3}{*}{LNK}} & 1st & $0.984^{(4)}$ & 1st & $0.982^{(4)}$ & 1st & $0.993^{(1)}$ \\
  &  & 2nd & $0.977^{(1)}$ & 2nd & $0.971^{(1)}$ & \textbf{2nd} & \textbf{0.990} \\
   &  & \textbf{3rd} & \textbf{0.972} & \textbf{3rd} & \textbf{0.960} & 3rd & $0.988^{(4)}$  \\
\hline
\multicolumn{2}{c}{\multirow{4}{*}{BIO}} & 1st & $0.862^{(4)}$ & 1st & $0.862^{(4)}$ & 1st & $0.955^{(1)}$  \\
  &  & 2nd & $0.794^{(1)}$ & 2nd & $0.782^{(1)}$ & \textbf{2nd} & \textbf{0.915} \\
  &   & 3rd & $0.752^{(2)}$ & \textbf{3rd} & \textbf{0.667} & 3rd & $0.890^{(5)}$ \\
   &   & \textbf{5th} & \textbf{0.699} &  &  &  &  \\
\hline
\hline
\multicolumn{2}{c}{Dataset} & \multicolumn{2}{c}{Fluo-N2DH-GOWT1} & \multicolumn{2}{c}{PhC-C2DL-PSC} & \multicolumn{2}{c}{Fluo-C2DL-MSC} \\
\hline

\multicolumn{2}{c}{\multirow{3}{*}{LNK}} & 1st & $0.986^{(4)}$ & 1st & $0.992^{(4)}$ & 1st & $0.947^{(4)}$  \\
  &  & \textbf{2nd} & \textbf{0.982} & 2nd & $0.991^{(1)}$ & \textbf{2nd} & \textbf{0.924}   \\
   &  & 3rd & $0.978^{(1)}$ & \textbf{3rd} & \textbf{0.991} & 3rd & $0.921^{(1)}$ \\
\hline
\multicolumn{2}{c}{\multirow{4}{*}{BIO}} & 1st & $0.788^{(4)}$ & 1st & $0.862^{(4)}$ & 1st & $0.727^{(4)}$ \\
  &  & 2nd & $0.723^{(2)}$ & 2nd & $0.812^{(1)}$ & 2nd & $0.657^{(1)}$ \\
  &   & 3rd & $0.718^{(1)}$ & 3rd & $0.805^{(2)}$ & 3rd & $0.647^{(3)}$  \\
   &   & \textbf{6th} & \textbf{0.661} & \textbf{5th} & \textbf{0.792} & \textbf{6th} & \textbf{0.592} \\
\hline
\end{tabular}
\end{table}

\begin{figure}
\includegraphics[width=\textwidth]{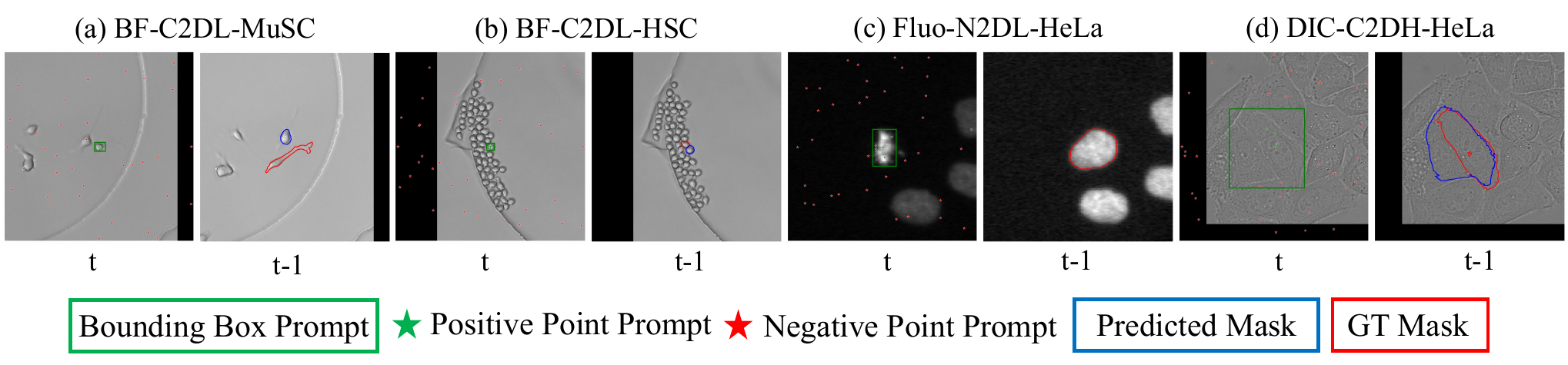}
\caption{Visualization of several failed cases of \gls{sam2} prediction results. Left: input image patch at time \( t \) with prompts. Right: output image patch at time \( t-1 \) with the predicted mask and the ground truth mask. The depicted examples are from CTC~\cite{ctc}. (a) and (c) show cases where the appearance of the cell changes significantly. In (c), the network fails to predict the mask. (b) illustrates a challenging situation with very high cell density. (d) shows that the bounding box might include neighboring cells, causing the neighboring cell to be incorrectly added.} \label{fig:visualize_err}
\end{figure}

Among the competing methods, KTH-SE (2)~\cite{kth-se} employs a global linking algorithm and optimizes parameters using a coordinate ascent approach with ground truth data. EPFL-CH (1)~\cite{trackastra} is a fully-supervised transformer-based network trained with tracking annotations. SIAT-CN (5) links cells using Euclidean distance and neighborhood similarity, though the exact similarity metrics were not given at the time of submission. PAST-FR (4) is an unsupervised approach based on optical flow enhanced Kalman filtering, while MON-AU (3) is also unsupervised and relies on flow network optimization. As a result, our approach achieves the top 3 in LNK score on average across all 13 datasets. Specifically, it achieves top 3 performance in 6 datasets for LNK and in 3 datasets for BIO. As a fully-unsupervised approach, our method demonstrates high performance and generalizability based on the LNK measurement, which directly evaluates linking quality. However, the BIO score is lower since our method focuses on consecutive time frames and local patches, improving efficiency but limiting the global overview of entire trajectories and cell relationships over larger areas. This can lead to breakpoints or ID switches in long trajectories, reducing long-term accuracy, which is more heavily weighted in the BIO score. Fig.~\ref{fig:visualize_err} shows several failed cases of the linking algorithm.

\subsection{Large-scale 3D Cell Tracking and Segmentation Results}
We evaluate our tracking and segmentation algorithm on two large-scale datasets from \gls{ctc}, each containing up to around 5000 cells. Since our approach is fully-unsupervised, meaning no ground truth information from the training set is used, we apply it directly to the training set without parameter fine-tuning and evaluate its performance using ground truth annotations. For comparison, we benchmark our method against the leading approach in the current \gls{ctc} rankings, as summarized in Tab.~\ref{tab2}. It is important to note that this is not a direct comparison, as the competitive methods are evaluated on the challenge's official test set, while our results are based on the training set. However, based on our linking algorithm's performance, we observe that training and test set scores are highly similar in our case, since our method is unsupervised and does not overfit the training set. In the supplementary materials (Suppl. Video Part 2,3,4,5), the results for both training and test sequences are provided. It can be observed that the quality of both results is at the same level. The fine-tuning and inference were done using the same hardware mentioned previously, with SAM-Med3D-turbo employed as the pre-trained weights for segmentation.

\begin{table}[ht]
\caption{Tracking and segmentation results on the large-scale datasets from \gls{ctc}~\cite{ctc}; the number in the superscript represents the corresponding method; our submission is in bold.}\label{tab2}
\centering
\begin{tabular}{cc cc cc cc}
\hline
\multicolumn{2}{c}{Dataset} & \multicolumn{2}{c}{Average Score} & \multicolumn{2}{c}{Fluo-N3DL-TRIC} & \multicolumn{2}{c}{Fluo-N3DL-TRIF}  \\
\hline
\multicolumn{2}{c}{\multirow{4}{*}{SEG}} & 1st & $0.714^{(6)}$ & 1st & $0.791^{(6)}$ & 1st & $0.746^{(9)}$  \\
 && 2nd & $0.700^{(9)}$ & 2nd & $0.766^{(8)}$ & \textbf{2nd} & \textbf{0.695} \\
  && \textbf{3rd} & \textbf{0.682} & 3rd & $0.680^{(7)}$ & 3rd & $0.684^{(10)}$ \\
  && 4th & $0.670^{(8)}$ & \textbf{4th} & \textbf{0.668} & 4th & $0.654^{(7)}$\\
\hline
\multicolumn{2}{c}{\multirow{4}{*}{TRA}}& 1st & $0.954^{(7)}$ & 1st & $0.952^{(7)}$ & 1st & $0.955^{(7)}$ \\
 & & \textbf{2nd} &  \textbf{0.910} & 2nd & $0.942^{(6)}$ & 2nd & $0.936^{(9)}$\\
 & & 3rd & $0.908^{(6)}$  &  \textbf{3rd} & \textbf{0.899} & \textbf{3rd} & \textbf{0.921}\\
 &  & 4th & $0.895^{(9)}$  & 4th & $0.854^{(9)}$ & 4th & $0.886^{(10)}$\\
\hline
\end{tabular}
\end{table}

Among the listed algorithms, KTH-SE (6)~\cite{kth-se} employs the Viterbi algorithm for linking and utilizes ground truth data to train a multinomial logistic regression classifier for cell detection. RWTH-GE (10)~\cite{twang} applies Laplacian of Gaussian-based cell detection with nearest-neighbor linking to generate the full tracklets. MPI-GE-CBG (7) is a supervised method, though its exact methodology is not publicly available. CZB-US (9)~\cite{ultrack} uses a difference of Gaussians filter to detect cells and utilizes \gls{ilp} for linking. Its hyperparameters are optimized via grid search and cross-validation on the training set. KIT-GE (8)~\cite{embedtrack} performs tracking through displacement estimation and trains a CNN-based network for segmentation, which is supervised. As a result, our approach achieves the third place in SEG score and second place in TRA score by averaging the result from both datasets, even outperforming some of the supervised methods.

\section{Discussion}
In this work, we present a foundation-model-based \textbf{zero-shot} cell segmentation and tracking pipeline for 2D and 3D time-lapse microscopy image sequences, aiming to reconstruct full cell lineage graphs with detailed migration and mitosis analysis. We demonstrate that pre-trained foundation models can be applied to solve microscopy-specific challenges without requiring extensive fine-tuning. We incorporate \gls{sam2} into our processing pipeline to reconstruct full tracklets from an incomplete set of pre-segmented masks, evaluating its performance across 13 different datasets. As a fully-unsupervised approach, our method achieves results comparable to state-of-the-art supervised techniques while exhibiting strong generalizability across diverse data types. Furthermore, we extend our framework to large-scale 3D+t datasets containing thousands of cells. By leveraging \gls{sam2} and SAM-Med3D in combination with automatically generated rough detections and masks, our method efficiently segments and tracks cells while maintaining accuracy close to leading approaches.

The current approach is limited when the dataset has huge spatial movement or very large time intervals. In future work, we plan to solve these challenges by integrating motion prediction of the whole structure movement between frames. Furthermore, integrating life cycle prediction models could provide deeper insights into cell behavior, improving the biologically inspired measures of the reconstructed lineage graphs.

\begin{credits}
\subsubsection{\ackname} This project was supported by Deutsche Forschungsgemeinschaft DFG (ZC, Grant Number STE 2802/5-1).

\subsubsection{\discintname}
The authors have no competing interests to declare that are relevant to the content of this article.
\end{credits}

%
%
%
\bibliographystyle{splncs04}
\bibliography{sam4cell-tracking}
\end{document}